\documentclass[12pt,oneside,reqno]{amsart}

\usepackage{graphicx}
\usepackage{pict2e}
\usepackage{amssymb}
\usepackage{amsthm}                
\usepackage[ margin={2cm,3cm}, headheight=3cm]{geometry}  
\usepackage{graphics}
\usepackage{epstopdf}
\usepackage{amsmath}
\usepackage{graphicx}
\usepackage{subfigure}
\usepackage{latexsym}
\usepackage{amsmath}
\usepackage{amsfonts}
\usepackage{amssymb}
\usepackage{tikz}
\usepackage{mathdots}
\usepackage{comment}
\usepackage{float}
\usepackage[mathscr]{euscript}
\usepackage{enumerate}
\usepackage{epsfig}
\usepackage { graphicx }
\usepackage { hyperref }
\usepackage { graphics }
\usepackage { graphicx }
\usepackage{MnSymbol} 
\usepackage{booktabs}
\usepackage{wrapfig,lipsum,booktabs}

\usepackage{caption}

\usepackage{algorithm}

\usepackage{etoolbox}
\usepackage{fancyhdr}

\usepackage {eufrak}
\usepackage[utf8]{inputenc}
\usepackage{longtable}
\usepackage[lite]{amsrefs}

\theoremstyle{definition}

\theoremstyle{definition}

\theoremstyle{remark}

\numberwithin{equation}{section}

\usepackage{newunicodechar}
\numberwithin{equation}{section}

\DeclareUnicodeCharacter{00A0}{~}
\title{Cell Complex Neural Networks}
\author{Mustafa Hajij}
\address{Santa Clara University}
\email{mhajij@scu.edu}

\author{Kyle Istvan}
\email{kyleistvan@gmail.edu}

\author{Ghada Zamzmi}
\address{University of South Florida}
\email{ghadh@mail.usf.edu}

\date{}
\keywords{}
\dedicatory{}

\begin{document}

\maketitle
\thispagestyle{fancy}%
\begin{abstract}
  Cell complexes are topological spaces constructed from simple blocks called cells. They generalize graphs, simplicial complexes, and polyhedral complexes that form important domains for practical applications. They also provide a combinatorial formalism that allows the inclusion of complicated relationships of restrictive structures such as graphs and meshes. In this paper, we propose \textbf{Cell Complexes Neural Networks (CXNs)}, a general, combinatorial and unifying construction for performing neural network-type computations on cell complexes. We introduce an inter-cellular message passing scheme on cell complexes that takes the topology of the underlying space into account and generalizes message passing scheme to graphs. Finally, we introduce a unified cell complex encoder-decoder framework that enables learning representation of cells for a given complex inside the Euclidean spaces. In particular, we show how our cell complex autoencoder construction can give, in the special case \textbf{cell2vec}, a generalization for node2vec.
\end{abstract}

\section{Introduction}
Motivated by the recent success of neural networks in various domains and data types (e.g., \cite{krizhevsky2012imagenet,lecun1998gradient,simonyan2014very,szegedy2015going,sutskever2014sequence,bahdanau2014neural}), we propose \textit{Cell Complex Neural Networks (CXNs)}, a general unifying scheme that allows neural network-type computations on cell complexes; i.e., we define a neural network structure on cell complexes. 

Cell complexes are topological spaces constructed from pieces called \textit{cells} that are homeomorphic to topological balls of varying dimensions. They form a natural generalization of graphs, simplicial complexes, and polyhedral complexes \cite{hatcher2005algebraic}. They also provide a combinatorial formalism that allows the inclusion of complicated relationships not available to more restrictive structures such as graphs and simplicial complexes, while retaining most of intuitive structure of these simpler objects.



Because the simplest nontrivial types of cell complexes are graphs \cite{hatcher2005algebraic}, our work can be considered as a generalization of Graph Neural Networks (GNNs) \cite{bruna2013spectral}. Earlier work that generalizes regular Convolutional Neural Network (CNN) to graphs was presented in \cite{gori2005new} and extended in \cite{scarselli2008graph,gallicchio2010graph,kipf2016semi}. Further, a significant effort has been made towards generalizing deep learning to manifolds, most notably geometric deep learning and the work of Bronstein et al. \cite{bronstein2017geometric,boscaini2015learning}. Other work includes utilizing filters on local patches using geodesic polar coordinates \cite{masci2015geodesic} and heat kernels propagation schemes \cite{boscaini2016learning}, among many others \cite{kokkinos2012intrinsic,boscaini2016learning,shuman2016vertex,wu20153d,monti2017geometric}. We refer the reader to recent reviews \cite{zhou2018graph,wu2020comprehensive} of GNNs and its variations and to \cite{cao2020comprehensive} for a recent survey on geometric deep learning.

The main contributions of this work are summarized as follows. We propose \textit{CXNs}, a general unifying and simple training scheme on cell complexes that vastly expands the domains upon which we can apply deep learning protocols. Our method encompasses most of the popular types of GNNs, and generalizes those architectures to higher-dimensional domains such as 3D meshes, simplicial complexes, and polygonal complexes. The training on a cell complex is defined in an entirely combinatorial fashion, and hence, naturally extends general message passing schemes currently utilized by GNNs. This combinatorial description allows for intuitive manipulation, conceptualization, and implementation.

We introduce an \textit{inter-cellular message passing scheme} on cell complexes that takes the topology of the underlying space into account. Precisely, we define a message passing scheme that is induced from the boundary and coboundary maps used to compute homology and cohomology in algebraic topology. As a concrete example of this scheme, we define \textit{Convolutional Cell Complex Networks (CCXN)}. Also, we propose a \textit{Cell Complex Autoencoder (CXNA)} to incorporate CXNs in a deep learning model and meaningfully represent cells of the complex in the Euclidean space. We provide examples of representational learning on cell complexes that generalize well-known representational learning on graphs such as graph factorization \cite{angles2008survey} and node2vec \cite{angles2008survey}. Computationally, a cell complex net is defined using adjacency matrices, analogous to those used to encode the structure of a graph neural network. This means their implementation can be readily adapted from the existing graph neural networks libraries (e.g., \cite{fey2019fast}).

The rest of this paper is organized as follows. Section \ref{sec.cell_complex} presents background and important notations about cell complexes. The proposed \textit{CXNs} is presented in Section \ref{cell complex nets}. Cell complex autoencoder \textit{CXNA} is introduced in Section \ref{autoencoder}. The topology-based message passing scheme on CXNs is discussed in Section \ref{general} of the Appendix followed by a presentation of potential applications in Section \ref{potenial}. 

\subsection{Cell Complexes: Background and Notations}
\label{sec.cell_complex}





\begin{figure}[h]
  \centering
   {\includegraphics[scale=0.11]{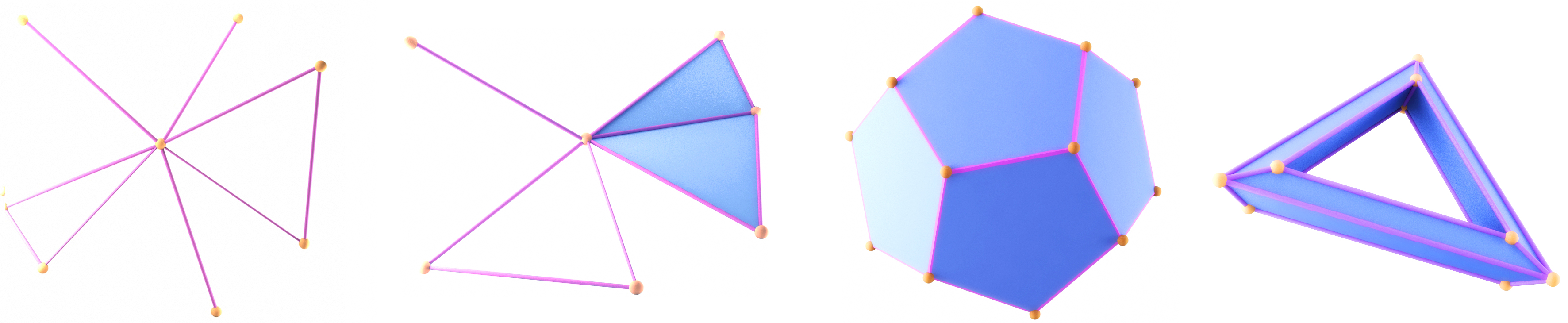}
    \caption{Examples of cell complexes.}
  \label{examples}}
\end{figure}

A cell complex is a topological space $X$ obtained as a disjoint union of \textit{cells}, each of these cells is homeomorphic to the interior of a $k$-Euclidean ball for some $k$. These cells are attached together via attaching maps in a locally reasonable manner\footnote{The reader is referred to \cite{hatcher2005algebraic} for further technical details of cell complex definition and algebraic topology.}. In our setting, the set of $k$-cells in $X$ is denoted by $X^k$, and it is called the $k-$ \textit{skeleton} of $X$. The set of all cells in $X$ whose dimension is less than $k$ is denoted by $X^{<k}$. The set $X^{>k}$ is defined similarly. The dimension of a cell $c\in X$ is denoted by $d(c)$, and the dimension of a cell complex is the largest dimension of one of its cells. See Figure \ref{examples} for various examples of cell complexes. 

A cell complex is called \textit{regular} if every attaching map is a homeomorphism onto the closure of the image of its corresponding cell. In this paper, all cell complexes will be regular and consist of finitely many cells. Regular cell complexes generalize graphs, simplicial complexes, and polyhedral complexes while retaining many desirable combinatorial and intuitive properties of these simpler structures. The information of attaching maps of a regular cell-complex are stored combinatorially in a sequence
of matrices called the boundary matrices ($\partial_k : \mathbb{R}^{|X^{k}|}  \to \mathbb{R}^{|X^{k-1}|}$). These matrices describe, roughly speaking, the number of times $k$-cells wrap around $(k-1)$-cells in $X$. The definition of these matrices $\partial_k$ depends on whether the cells of $X$ are oriented or not. Our method is applicable to both oriented and non-oriented cell complexes. However, we only discuss the non-oriented case for the sake of brevity, and leave the oriented case to section 4.3 of the Appendix. Since our cell complexes are regular and non-oriented, the entries of $\partial_k$ are in $\{0,1\}$. Dually, we define $\partial^*_k:\mathbb{R}^{|X^{k-1}|} \to \mathbb{R}^{|X^{k}|} $ to be the transpose of the matrix $\partial_k$.

Let $X$ be a cell complex, $c^n$ denotes an $n$-cell in $X$, and $facets(c^n)$ denotes the set of all $(n-1)$-cells $X$ that are incident to $c^n$. Similarly, let $cofacets(c^n)$ denotes the set of all $(n+1)$-cells of $X$ that are incident to $c^n$. Note that $facets(c^n)$ or $cofacets(c^n)$ might be the empty set. We say that two $n$-cells $a^n$ and $b^n$ are \textit{adjacent} if there exists an $(n+1)$-cell $c^{n+1}$ such that $a^n,b^n \in facets(c^{n+1})$. Likewise, we say that $a^n$ and $b^n$ are \textit{coadjacent} in $X$ if there exists an $(n-1)$-cell $c^{n-1}$ with  $a^n,b^n \in cofacets(c^{n-1})$. The set of all cells adjacent to a cell $a$ in $X$ is denoted by $\mathcal{N}_{adj}(a)$ while the set of all cells coadjacent to a cell $a$ in $X$ is denoted by $\mathcal{N}_{co}(a)$. If $a^n,b^n $ are $n$-cells in $X$, then we define the set $\mathcal{CO}[a^{n}, b^{n} ]$ to be the intersection of $cofacets(a^n) \cap cofacets(b^n)$. Note that this notation is symmetric:  $\mathcal{CO}[a^{n}, b^{n} ]=\mathcal{CO}[b^{n}, a^{n} ]$ \footnote{This is not the case when $X$ is not oriented. See Section \ref{sm} in the Appendix for more details.}. Similarly, the set $\mathcal{C}[a^{n}, b^{n} ]$ is defined to be the intersection of $facets(a^n) \cap facets(b^n)$. 
  
\begin{figure}
    \centering
  \includegraphics[width=7.4cm]{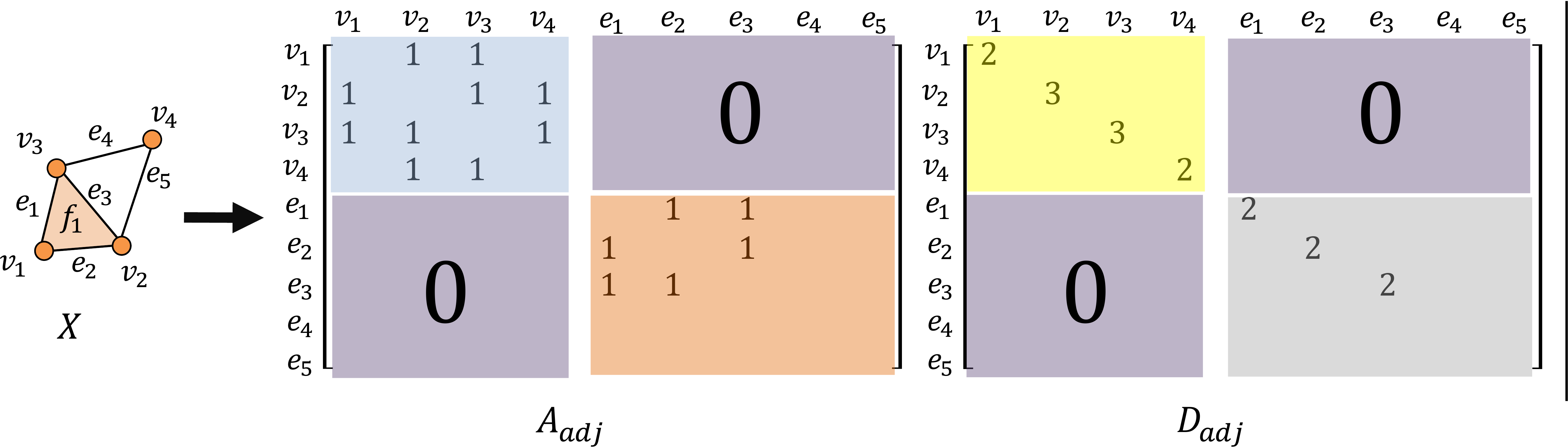} 
   \includegraphics[width=7.4cm]{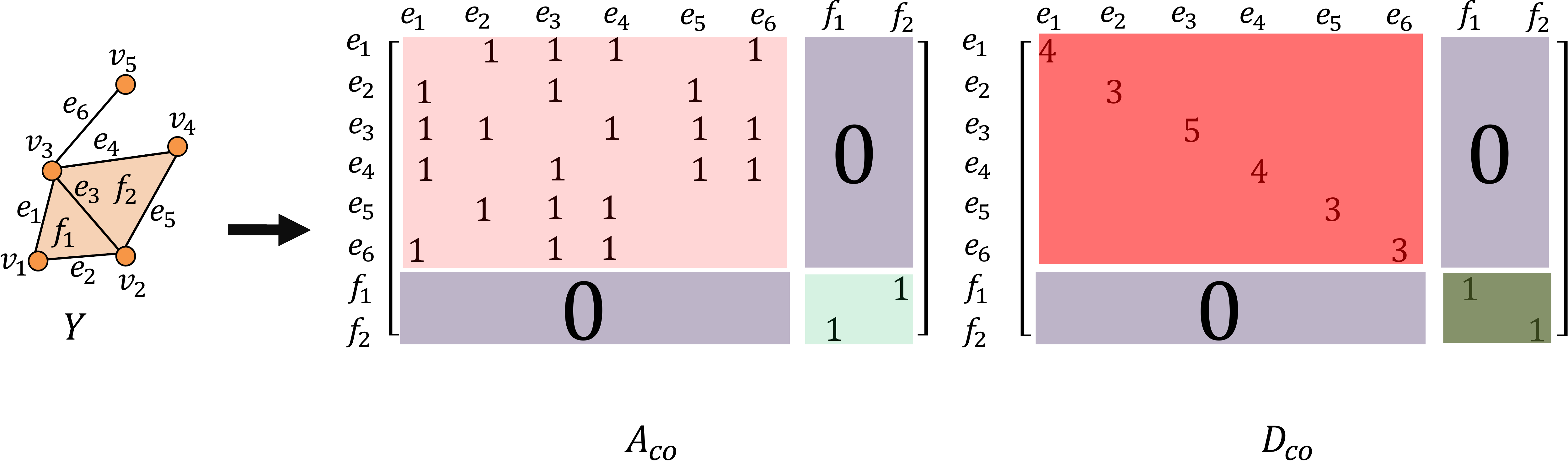} 
    \caption{Examples of adjacency and coadjacency matrices for simplicial complexes. Left: the adjacency matrix $A_{adj}$ and adjacency degree matrix $D_{adj}$ of the simplicial complex $X$. The blue and the orange submatrices in  $A_{adj}$ represent $A^0_{adj}$ and $A^1_{adj}$ of $X$, respectively. The yellow and grey submatrices of $D_{adj}$ represent $D^0_{adj}$ and $D^1_{adj}$, respectively. Right: the coadjacency matrix $A_{co}$ and coadjacency degree matrix $D_{co}$ of the simplicial complex $Y$. The pink and the light green matrices in $A_{co}$ represent $A^1_{co}$ and $A^2_{co}$ of $Y$, respectively. The red and dark green submatrices of $D_{co}$ represent $D^1_{co}$ and $D^2_{co}$, respectively. Note that the entries of adjacency and coadjacency matrices in case of a simplicial complex, and more generally polyhedral complexes, are always in $\{0,1\}$.}
    \label{adjcoadj}
\end{figure}

  Note that these notions generalize the analogous notions of adjacency and coadjacency matrices defined on graphs. More precisely, let $X$ be a cell complex of dimension $n$. Let $N$  denotes the total number of cells in the complex $X$ and define $\hat{N}:=N-|X^{n}|$. Let $c_1,\cdots , c_{\hat{N}}$ denotes all the cells in $X^{<n}$. Then, we define the matrix $A_{adj}$ of dimension $\hat{N} \times \hat{N}$ by storing a $|\mathcal{CO}[c_i, c_j]|$ in $A_{adj}(i,j)$ if the cell $c_i$ is adjacent to $c_j$, otherwise, we store a $0$ in $A_{adj}(i,j)$. Note that the matrix $A_{adj}(i,j)$ does not store the adjacency information of $n$-cells in $X$ since the dimension of the complex $X$ is $n$. We denote the adjacency matrix between $k$-cells in $X$ by $A^k_{adj}$, where $ 0 \leq  k< n$. The \textit{adjacency degree matrix} $D_{adj}$ is defined via $D_{adj}(i,i)=\sum_{j}A_{adj}(i,j)$, and hence, we define the \textit{adjacency degree matrix between $k$-cells} $D_{adj}^k$ ($0 \leq k <n$) similarly. Finally, the coadjacency matrix $A_{co}$, the coadjacency degree matrix $D_{co}$, the $k$-cells coadjacency matrices $A^k_{co}$, and the $k$-cells coadjacency degree matrices $D^k_{co}$ for $ 0 < k \leq n $, are defined similarly. Examples of these matrices are presented in Figure \ref{adjcoadj}. 
  
 \vspace{-5pt}

\vspace{-5pt}

\section{Cell Complex Neural Networks (CXNs)}
We define below a general CXNs using a message passing scheme that generalizes the notations of message passing schemes in graphs. Section 4 of the Appendix provides a brief review of message passing schemes on graphs (Section \ref{gnn}), and introduce message passing schemes on cell complexes (Section \ref{sm}.4). We also present here \textit{CCXN} (Section \ref{CCNX}) as an example of CXNs. The forward propagation computation of a cell complex neural net requires the following data as inputs: (1) A cell complex $X$ of dimension $n$, possibly oriented and (2) For each $m$-cell $c^m$ in $X$, we have an initial vector $h_{c^m}^{(0)} \in \mathbb{R}^{l^0_m}$. The forward propagation algorithm then performs a sequence of message passing executed between cells in $X$. Precisely, given the desired depth $L>0$ of the net one wants to define on the complex $X$, the forward propagation algorithm on $X$ consists of $L\times n $ \textit{inter-cellular message passing scheme} defined for $0 <  k \leq L $:
    
      \begin{equation}
      \label{intercell}
      \small{
          h_{c^0}^{ (k) }:= \alpha^{(k)}_0 \bigg( h_{c^0}^{ (k-1) }, E_{ a^0 \in \mathcal{N}_{adj}(c^0)}\Big( \phi^{(k)}_0 ( h_{c^0}^{(k-1)},h_{a^0}^{(k-1)}, F_{e^1 \in \mathcal{CO}[a^0,c^0]  } ( h_{e^1}^{(k-1))} ) \Big) \bigg)  \in \mathbb{R}^{l^k_0},  }
      \end{equation}

             $\vdots$
      \begin{equation}
      \label{second}
      \small{
             h_{c^{n-1}}^{ (k) }:= \alpha^{(k)}_{n-1} \bigg( h_{c^{n-1}}^{ (k-1) }, E_{ a^{n-1} \in \mathcal{N}_{adj}(c^{n-1})}\Big( \phi^{(k)}_{n-1} ( h_{c^{n-1}}^{(k-1)},h_{a^{n-1}}^{(k-1)}, F_{e^{n} \in \mathcal{CO} [a^{n-1},c^{n-1}]  } ( h_{e^n}^{(0)}) \Big) \bigg) \in\mathbb{R}^{l^k_{n-1}}}
      \end{equation}
\vspace{-5pt}
where $h_{ e^m}^{(k)}, h_{ a^m}^{(k)},h_{ c^m}^{(k)} \in \mathbb{R}^{l_{m}^k}$, $E,F$ are permutation invariant differentiable functions\footnote{The permutation invariance condition on $F$ and $E$ may not be necessary in general (e.g. on triangulated meshes only cyclic invariance is needed).  However, a permutation invariant function is needed whenever there is no canonical way to order the cofaces adjacent to a face in the complex. }, and $\alpha^{(k)}_j,\phi^{(k)}_j$ are trainable differentiable functions where, $0 \leq j \leq n-1 $ and $0 < k \leq L $ \footnote{Examples of the functions  $\alpha^{(k)}_j,\phi^{(k)}_j$ in practice are MLP. A concrete example is given in \ref{CCNX}}.        
Note that for each cell $a^{n}$ in $X$, the vectors $h_{a^n}^{(0)}$ are never updated during the training of a CXN. Although the formulation above is simple, it can generalize most types of the popular types of GNNs (e.g., \cite{kipf2016semi,velivckovic2017graph} ). 


\label{cell complex nets}


\begin{figure}[t]
  \centering
   {\includegraphics[scale=0.035]{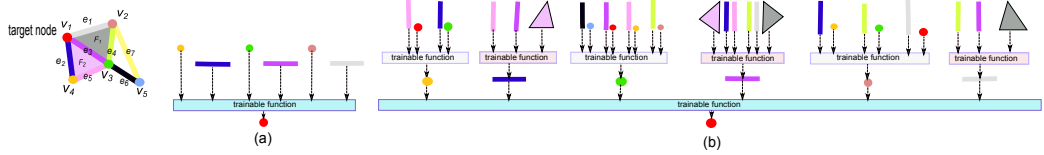}
    \caption{Two layers Cell Complex Neural Network (CXN). The example demonstrates a simplicial complex neural network for clarity.  The computation is demonstrated with respect to the red target vertex. The information flow goes from lower cells to higher incident cells. }
  \label{DMF1}}
\end{figure}

Figure \ref{DMF1} demonstrates the above construction/formulation on a simplicial complex network with depth $L=2$. We will use $X$ to denote this complex. Note that we abuse the notation in the figure and do not distinguish between a simplex $s$ and its vector $h^{(k)}_s$. For each vertex $\{v_i\}_{i=1}^5$ in $X$, we assume we are given a vector $h^{(0)}_{v_i}$; we have $h^{(0)}_{e_j}$ for each edge $\{e_{j}\}_{j=1}^7$, and have $h^{(0)}_{F_i}$ for the faces $\{F_i\}_{i=1}^2$. 

In the first stage, we start the computation for cells with dimension $0$. In this stage, each $v_i$, $0\leq i\leq 5$, computes : 
             $
          h_{v_i}^{ (1) }:= \alpha^{(k)}_0 \bigg( h_{v_i}^{ 0 }, E_{ v_j \in \mathcal{N}_{adj}(v_i)}\Big( \phi^{(1)}_0 ( h_{v_i}^{(0)},h_{v_j}^{(0)}, h_{e_{ij}}^{(0)} ) \Big) \bigg) 
       $, where $e_{ij}$ is the edge that connects $v_i$ to $v_j$. Figure \ref{DMF1} (a) shows this graph for $v_1$. Notice that all nodes share the same trainable functions $\alpha^{(1)}_0$ and $\phi^{(1)}_0$.
Further, each edge $e_i$ induces a computational graph and computes $h^{(1)}_{e_i}$, $1 \leq i \leq 7$: 
           $
          h_{e_i}^{ (1) }:= \alpha^{(k)}_1 \bigg( h_{e_i}^{ 0 }, E_{ e_j \in \mathcal{N}_{adj}(e_i)}\Big( \phi^{(1)}_1 ( h_{e_i}^{(0)},h_{e_j}^{(0)}, h_{F_{ij}}^{(0)} ) \Big) \bigg) $, here $ F_{ij}$ denotes the unique face that bounds both edges $e_i$ and $e_j$. Note that all edges share the same trainable functions $\alpha^{(1)}_1$ and $\phi^{(1)}_1$.
      In stage $2$, we compute $h^{(2)}_{s}$ for all simplices $s$ of dimension 0 and 1. Figure \ref{DMF1} shows this computation for $h^{(2)}_{v_1}$. Note that (1) the computational graphs that feed into it are the ones computed in stage 1 and (2) how the information from this node flows from the surrounding nodes, edges, and faces.

\subsection{Convolutional Cell Complex Nets (CCXN)}
\label{CCNX}
We present \textit{CCXN}, the simplest type of cell complex neural networks. Specifically, using the adjacency matrices on a cell complex $X$ defined in \ref{sec.cell_complex} and Figure \ref{adjcoadj}, we extend the definition of convolutional graph neural networks (\textit{CGNN}) \cite{kipf2016semi} to \textit{CCXN}. The input for a CCXN is specified by cell embeddings $H^{(0)} \in \mathbb{R}^{ \hat{N} \times d}$ that define the initial cells features on every cell in $X^{<n}$. Here, $d$ is the embedding dimension of the cells. The \textit{convolutional message passing scheme} on $X$ is defined by :
\begin{equation}
\label{CCXN}
    H^{(k)}:=\text{ReLU}( \hat{A}_{adj} H^{(k-1)} W^{(k-1)})
\end{equation}
where $\hat{A}_{adj}=I_{\hat{N}} + D^{-1/2}_{adj}{A}_{adj} D^{-1/2}_{adj},   $ $H^{(k)} \in \mathbb{R}^{\hat{N}\times d}  $
 are the cell embeddings computed after $k$ steps of applying \ref{CCXN}, and $W^{(k-1)} \in \mathbb{R}^{
d\times d}$ is a trainable weight matrix at the layer $k$. We discuss the following few remarks about the definition above. First, observe that we chose the embedding dimension of cells to be $d$ for all $H^{(k)}$. However, this restriction is not necessary in general and we chose it only for notational convenience. Second, we train the CCXN with a single weight $W^{k}$ for every layer $k$ for all cells. This restriction is also not necessary in general. As indicated in equations (\ref{intercell},\ref{second}), one may choose to train a different CCXN for every $k$-cells adjacency matrix $A_{adj}^k$ individually. In this case, we need to train $n-1$ cell complex networks. Finally, the matrix $\hat{A}_{adj}$ in equation \ref{CCXN} is typically normalized to avoid numerical instabilities
when stacking multiple layers \cite{kipf2016semi}. Specifically, with the \textit{renormalization trick}, we make the substitution $I_{\hat{N}} + D^{-1/2}_{adj}{A}_{adj} D^{-1/2}_{adj} \to  \Tilde{D}_{adj}^{-1/2}\Tilde{A}_{adj}\Tilde{D}_{adj}^{-1/2}$ in equation \ref{CCXN} where $\Tilde{A}_{adj}=A_{adj}+I_{\hat{N}}$ and $\Tilde{D}_{adj}(i,i)=\sum_{j}\Tilde{A}_{adj}(i,j)$. Observe that with this simplified case, this version for CCXN is a generalization of CGNN where the only difference being the generalized notion of adjacency on cell complexes.

\vspace{-8pt}
\section{Cell Complex Autoencoders (CXNA) and Cell Complex Representations}
\label{autoencoder}
In this section, we present how to incorporate the cell complex structure into a deep learning model. 
Given a cell complex $X$, we want to learn a function that embeds the cells of $X$ into some Euclidean space such that the structure information of these cells is preserved. Inspired by the success of graph autoencoders in representational learning on graphs \cite{hamilton2017representation}, we propose a general method to learn cell complex representations. While the method provided in \cite{hamilton2017representation} is general and encompasses various representational learning strategies (e.g., Graph Factorization \cite{angles2008survey}, node2vec \cite{grover2016node2vec}, and DeepWalk \cite{perozzi2014deepwalk}), it assumes a node to node message passing scheme using edges. Because our setting has to accommodate for different message passing schemes on a cell complex, we present below an autoencoder definition that is consistent with the inter-cellular message passing scheme in equation \ref{intercell}. Other cell autoencoder definitions that are consistent with different message passing schemes can be defined similarly. It is important to note that we are aware of a related work \cite{hacker2020k} appeared simultaneously with our work, and \cite{billings2019simplex2vec,schaub2020random} where a vector representation based on random walks on simplices in a simplicial complex was suggested. Contrary to these works, our approach is a unified framework that describes these special cases to a larger set of vector-based representations and a larger set of complexes.


A cell complex autoencoder consists of three components: (1) an encoder-decoder system, the trainable component of the autoencoder, (2) a similarity measure on the cell complex, which is a user-defined similarity function that represents a notion of similarity between the cells in the complex, and (3) a loss function, which is a user-defined function utilized to optimize the encoder-decoder system according to the similarity measure. Mathematically, let $X$ be a cell complex of dimension $n$. Then, an \textit{encoder} on $X$ is a function of the form:
$$
enc : X^{<n}\to \mathbb{R}^d.
$$
This encoder associates to every $k$-cell $c^k$ ($0 \leq k < n$) a feature vector $\textbf{z}_c \in \mathbb{R}^d$ that encodes the structure of this cell and its relationship to other cells in $X$. A \textit{decoder} is a function of the form: $$
    dec : \mathbb{R}^d \times \mathbb{R}^d\to \mathbb{R}^+
$$
This decoder associates to every pair of cell embeddings a measure of similarity that quantifies some notion of relationship between these cells. The pair $(enc,dec)$ on $X$ is called a \textit{cell complex encoder-decoder system} on $X$ \footnote{ In our definition of the encoder-decoder system, we chose to embed all cells on $X$ in the same ambient space. This assumption is not needed in general and we are only making this restriction for notational convenience. Alternatively, one may have a sequence of similarity measure to describe the similarity between cells that have the same dimension.}. The functions $enc$ and $dec$ are typically trainable functions that are optimized using user-defined similarity measure and loss function. A similarity measure on a cell complex is a function of the form $sim_X: X^{<n}\times X^{<n} \to \mathbb{R}^+$ such that $sim_X(a^k,c^{l})$ reflects a user-defined similarity between the two cells $a^{k}$ and $c^{l}$ in $X^{<n}$.  We will assume that $sim_X(a^{k},c^{l})=0$ whenever $k \neq l $. An example of a similarity measure defined on $X$ is $A_{adj}$ defined in Section \ref{CCNX}. We want the encoder-decoder system specified above to learn a representation embedding of the cells in $X^{<n}$ such that: $dec( enc(a^{k}), enc(c^l)) = dec( \textbf{z}_{a^k}, \textbf{z}_{c^l})\approx sim_{X}(a^{k},c^l).$ To this end, let $l : \mathbb{R} \times \mathbb{R}\to \mathbb{R} $ be a user-defined loss function and define: 
\begin{equation}
\label{loss}
    \mathcal{L}_k=\sum_{\textit{all possible }  \mathcal{CO}[a^k,c^k] \subset X^{k+1}  } l(  dec(  enc(\textbf{z}_{a^{k}}), enc(\textbf{z}_{c^k})),sim(a^{k},c^k)),
    \end{equation}
and finally define $\mathcal{L}:=\sum_{k=0}^{n-1} \mathcal{L}_k$. 

Different concrete CXNAs can be provided as shown in Table \ref{tab:title}. After training the encoder-decoder model, we can use the encoder to generate the embeddings for $k$-cell, $0 \leq k <n $.

\begin{table}[h]
\centering
\caption{Various definitions of CXNAs. }
\label{tab:title} 
 \begin{tabular}{||c|c|c|c||} 
 \hline
Method &  Decoder & similarity & Loss \\ [0.5ex] 
 \hline
 Laplacian eigenmaps \cite{belkin2001laplacian}  &  $||\textbf{z}_{a}-\textbf{z}_{c}||_2^2$ & general & $dec(\textbf{z}_{a},\textbf{z}_{c}).sim(a,c) $ \\ 
 Inner product methods \cite{ahmed2013distributed} & $ \textbf{z}_{a}^T\textbf{z}_{c}$ & $A_{adj}(a,c)$ & $||  dec(\textbf{z}_{a},\textbf{z}_{c})-sim(a,c) ||_2^2$ \\
 Random walk methods \cite{grover2016node2vec,perozzi2014deepwalk} & $  \frac{
 e^{\textbf{z}_a^T \textbf{z}_c}}{\sum_{b\in X^{k}  } e^{\textbf{z}_a^T \textbf{z}_b } } $ & $p_X(a|c)$   & $  -log(dec(\textbf{z}_{a},\textbf{z}_{c})) $ \\
 [1ex] 
 \hline
 \end{tabular}
\end{table}
We end this section by noting how the random walk method given in Table \ref{tab:title} effectively defines \textit{cell2vec}, a cell complex representation method that generalizes node2vec \cite{grover2016node2vec} and DeepWalk \cite{perozzi2014deepwalk} to cell complexes\footnote{Random walks on graphs can be naturally extended to cell complexes by using the adjacency relations on cell complexes as a mean to define a random walk between cells of the same dimension.}. 



\bibliographystyle{abbrv}
\bibliography{refs_2}

\clearpage
\section{APPENDIX}
\label{sm}

\subsection{Why Cell Complex Nets?}
This work focuses on the construction of the necessary tools that perform neural networks-type computations over domains that have geometric and combinatorial characteristics. Our ultimate goal is to harness the power of deep learning in solving problems that arise when studying these domains or working on problems that are naturally modeled by such domains.

With the above goal in mind, one may wonder why we chose cell complexes for representation over other less general complexes available in algebraic topology? Cell complexes neural nets (CXNs) are generalization of graph neural networks (GNNs), and the gap of generalization between graphs and cell complexes contain many other complexes (e.g., simplicial complexes, polyhedral complexes, and $\Delta$-complexes) that are less general than cell complexes. We discuss below the reasons of why CXNs form a better and more expressive generalization than, say, graph and simplicial complex neural networks \footnote{We are aware of few related works about unoriented simplicial neural networks \cite{bunch2020simplicial,ebli2020simplicial} that were published at the same time our work got published. We note, however, that our approach is applicable to a more general set of complexes and handles both oriented and non-oriented cases. See Section \ref{oriented} for the oriented case.}. 


\begin{itemize}

\item \textbf{Hierarchical relational reasoning representation:}
Graphs are natural objects for modeling relations between entities. In this context and towards building intelligent behaviour, GNNs have been extensively explored to build relational reasoning between various objects \cite{santoro2017simple,zhang2020deep,chen2019graph,schlichtkrull2018modeling} and in knowledge graphs \cite{arora2020survey,wang2017knowledge}. However, we believe building intelligent behaviour goes beyond the ability to create relational reasoning between entities. Abstraction and analogy that humans are capable of require building relations among the relations in a hierarchical manner. The importance of this perspective is evident in the unprecedented success of deep learning models where complicated concepts are built from simpler ones in a hierarchical fashion. 

Therefore, utilizing GNNs for relational reasoning is ``shallow'' because the edges can be only used to model relationships between entities. Even labeled multi-graph \cite{schlichtkrull2018modeling,jung2020accurate} are insufficient to model hierarchical relational reasoning because the relationships modeled by the multi edges or multi nodes remain between the raw entities and no deeper relations can be made. CXNs have the ability to model hierarchical relational reasoning. Precisely, the 1-skeleton of a cell complex can represent the shallow relation between objects while the higher cells can be used to build more abstract relations between the relations in a hierarchical manner. Note that other objects, such as simplicial complexes, have uniform structure, in the sense that $k$-faces have fixed number of edges, which is not natural to model hierarchical complex relational reasoning. Within this context, knowledge graphs can be generalized to \textit{knowledge cell complexes} where entities and relations in knowledge graphs are replaced with hierarchical representation of abstract entities and relations between them.

\item \textbf{Nature of data and application:} Contrary to simplicial complex neural networks, our definition is a unifying and combinatorial framework that accommodates for almost all data forms that are significant in practice such as  polygonal 2d and 3d meshes. In addition to the nature of data, the application at hand determines, in many cases, the type of complexes one needs to work with. For instance, in several CAD \cite{piegl1996nurbs,marinov2006robust} and simulation applications \cite{yasseen2013sketch}, quad meshes are preferred over simplicial complexes \cite{bommes2013quad}. Quad meshes are also preferred when solving PDE on surface and are best suited for defining Catmull-Clark subdivision surfaces \cite{stam2003flows,halstead1993efficient}.


\item \textbf{Trainability of neural networks over geometric domains and resource efficiency:} As is well-known, GNNs are challenging to train for large graphs \cite{jia2020improving}, and simplicial complexes require massive amount of memory and computational power even without doing deep learning computations on them. Hence, it is essential to consider the complexity of the geometric object representation when designing a neural network over these domains. In this context, building geometric objects with cell complexes requires significantly less number of cells than building the same objects with simplicial complexes or other complexes. 

\end{itemize}

\subsection{Graph Neural Networks}
\label{gnn}
Given a graph $G=(V,E)$, a graph neural network on $G$ with depth $L>0$ updates a feature representation for every node in the graph $L$ times. Initially, every node $i$ is given a feature vector $h^{(0)}_i$. On the $k$ stage of the computation, each node $i$ in the graph collects messages from its neighbors $j$, represented by their feature vectors $h^{(k-1)}_j$, and aggregates them together to form a new feature representation $h^{(k)}_i$ for the node $i$. A graph neural network requires the following input data:  
 \begin{enumerate}
     \item A graph $G=(V,E)$.
     \item For each node $i \in V $ we have an initial vector $h_{i}^{(0)} \in \mathbb{R}^{l_0}.$
 \end{enumerate}

Given the above data, the feedforward neural algorithm on $G$ executes $L$ message passing schemes defined recursively for $0 \leq  k \leq L $ by:
      \begin{equation}
         h_{i}^{ (k) }:= \alpha^{k} \bigg( h_{i}^{ (k-1) }, E_{ j \in \mathcal{N}(i)}\Big( \phi^{k} ( h_{i}^{(k-1)},h_{j}^{(k-1)}, e_{i,j}    ) \Big) \bigg)  \in \mathbb{R}^{l_k},  
      \end{equation}
 
where $e_{ij} \in \mathbb{R}^D$ is an edge feature from the node $j$ to the node $i$, $E$ is a permutation invariant differentiable function, and $\alpha^{k},\phi^{k}$ are trainable differentiable functions. Note that at each stage $k$, all messages share the same differentiable functions $\phi^k$ and $\alpha^k$. 

\begin{figure}[H]
  \centering
   {\includegraphics[scale=0.13]{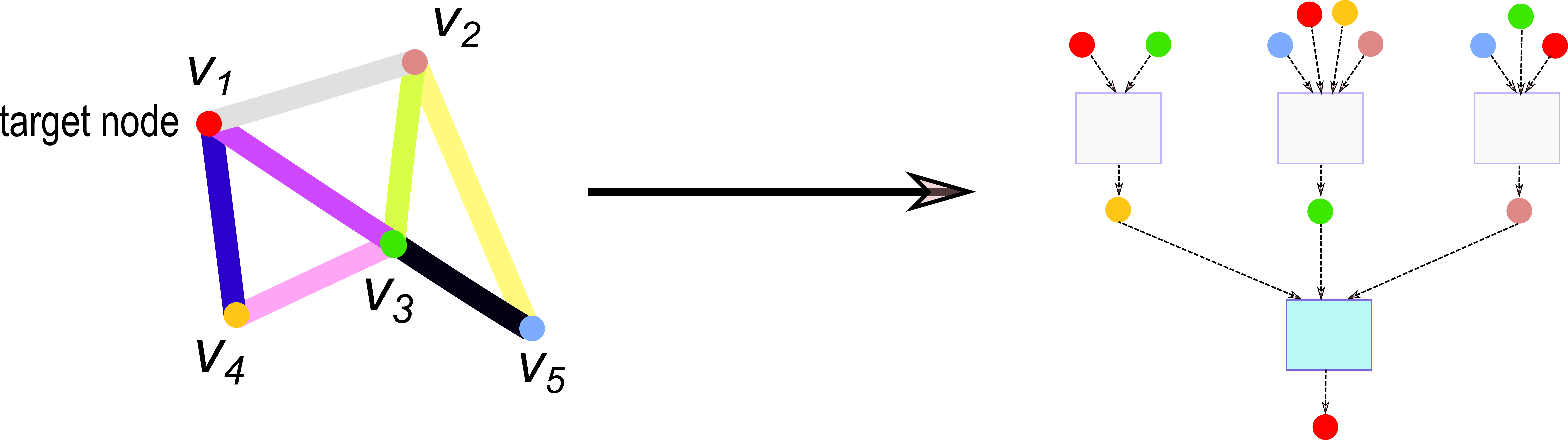}
    \caption{An example of graph neural net with depth $2$. The computation are only illustrated on the red node. In this figure, we abuse the notation and do not distinguish between a node $i$ and its associated vector $h^{(k)}_i$. The blue box represents the differentiable functions $\alpha^1$ and $\phi^1$ while the white box represents the functions $\alpha^2$ and $\phi^2$.  }
  \label{DMF}}
\end{figure}

Consider the graph given in Figure \ref{DMF}. Let's say we want to build a graph neural network on this graph with depth $2$. To illustrate the computation, we pick a vertex  $v_1$ in the graph. In the first stage, the surrounding neighbors of $v_1$, namely $\{v_2,v_3,v_4\}$, pass their messages to $v_1$. The information obtained from these messages are aggregated and combined together via trainable differentiable functions $\alpha^1$ and $\phi^1$. In the second stage, all neighbors of $v_1$ collect the messages information from their respective neighbors in a similar fashion as illustrated in Figure \ref{DMF}.

\subsection{The adjacency relation when complex $X$ is oriented}
\label{oriented}
\
In this section, we discuss the adjacency and coadjacency relations in a cell complex $X$ when $X$ is oriented. Recall that each cell $a$ in $X$ has two possible orientations. An \textit{oriented cell complex} is a cell complex in which every cell has a chosen orientation. When $X$ is regular, then the entries of $\partial_k$ are in $\{0,\pm 1\}$.

The definitions of $facets(c^n)$ and $cofacets(c^n)$ are more complicated for the oriented case as compared to the non-oriented case. When $X$ is oriented, we store along with each cell in $facets(c^n)$ and $cofacets(c^n)$ the orientation induced by $c_n$ with respect to the maps $\partial_n$ and $\partial^*_n$, respectively.


In this case, we use $facets^{+}(c^n) \subset facets(c^n) $, $facet^{-}(c^n) \subset facets(c^n) $ to the subsets of $facets(c^n)$ to denote the cells that are positively oriented and negatively oriented with respect to $c^n$, respectively. The set $cofacets^{+}(c^n) $ and the set $cofacets^{-}(c^n) $ are defined analogously. Observe that $facets(c^n)= facets^+(c^n) \cup facets^-(c^n) $ and $facets^+(c^n) \cap facets^-(c^n)= \emptyset$. Similarly, $cofacets(c^n)= cofacets^+(c^n) \cup cofacets^-(c^n) $ and $cofacets^+(c^n) \cap cofacets^-(c^n)= \emptyset $.

\begin{figure}[!t]
  \centering
   {\includegraphics[scale=0.25]{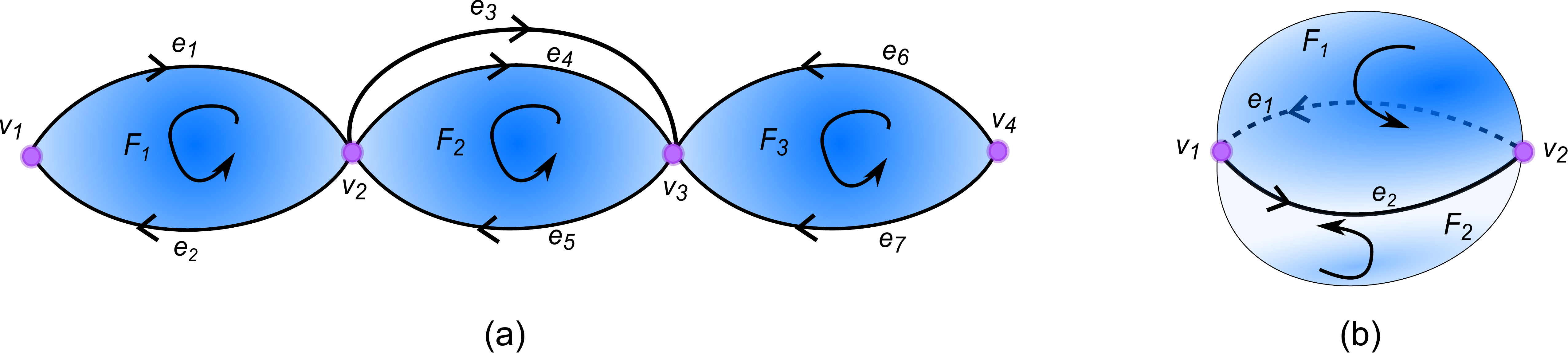}
    \caption{Examples of computing the adjacency and co-adjacency neighbors of cells in cell complexes.}
  \label{example}}
\end{figure}

Consider the cell complexes given in Figure \ref{example}, we compute few examples of the sets we define above to illustrate the concept. For Figure \ref{example} (a), we have  $cofacets(v_1)=\{-e_1,e_2\}$,
  $cofacets(v_2)=\{e_1,-e_2,-e_3,-e_4,e_5 \}$,   $cofacets(v_3)=\{e_3,e_4,-e_5,e_6,e_7\}$ and $cofacets(v_4)=\{-e_6,-e_7\}$. Moreover, $facets(F_1)=\{e_1,-e_2\}$. Finally, $cofacets(e_6)=\{F_3\}$ and $cofacets(e_7)=\{-F_3\}$. On the other hand, for Figure \ref{example} (b)  
$cofacets(e_1)=\{F_1,-F_2\}$ and $cofacets(e_2)=\{F_1,-F_2\}$.  

If $X$ is an oriented complex, then a cell $b^{n}$ is said to be \textit{adjacent} to an $n$-cell $a^{n}$ with respect to an $(n+1)$-cell $c^{n+1}$ when $a^{n} \in facets^+(c^{n+1}) $ and $b^n \in facets^-(c^{n+1}) $. Similarly,  an $n$-cell $b^{n}$ is said to be \textit{coadjacent} to $a^{n}$ with respect to an $(n-1)$-cell $c^{n-1}$ if $a^{n} \in cofacets^+(c^{n-1}) $ and $b^n \in cofacets^-(c^{n-1})$.  The set of all adjacent cells of a cell $a$ in $X$ is denoted by $\mathcal{N}_{adj}(a)$. Similarly, the set of all coadjacent cells of a cell $a$ in $X$ is denoted by $\mathcal{N}_{co}(a)$.

In Figure \ref{example} (a), we have $\mathcal{N}_{adj}(v_2)=\{v_1,v_3\}$ $\mathcal{N}_{adj}(v_4)=\emptyset$. For Figure \ref{example} (b), we have $\mathcal{N}_{co}(F_1)=\{F_2\}$. This is because $F_1 \in cofacets^+(e_1) $ and $F_2 \in cofacets^-(e_1) $. On the other hand,  $\mathcal{N}_{co}(F_2)=\emptyset$. Note that in this example $\mathcal{N}_{adj}(e_1)=\mathcal{N}_{adj}(e_2)=\emptyset$

If $a^n$ and $b^n $ are $n$-cells in $X$, then we define the set $\mathcal{CO}[a^{n}, b^{n} ]$ to be the intersection $cofacets(a^n) \cap cofacets^+(b^n)$. The set $\mathcal{CO}[a^{n}, b^{n} ]$ describes the set of all incident $(n+1)$-cells of $b^n$ that have $a^n$ as an adjacent cell. Note that in general $\mathcal{CO}[a^{n}, b^{n} ] \neq \mathcal{CO}[b^{n}, a^{n} ]$. Similarly, the set $\mathcal{C}[a^{n}, b^{n} ]$ is defined to be the intersection of $facets(a^n) \cap facets^+(b^n) $.

\begin{figure}[h]
  \centering
   {\includegraphics[scale=0.27]{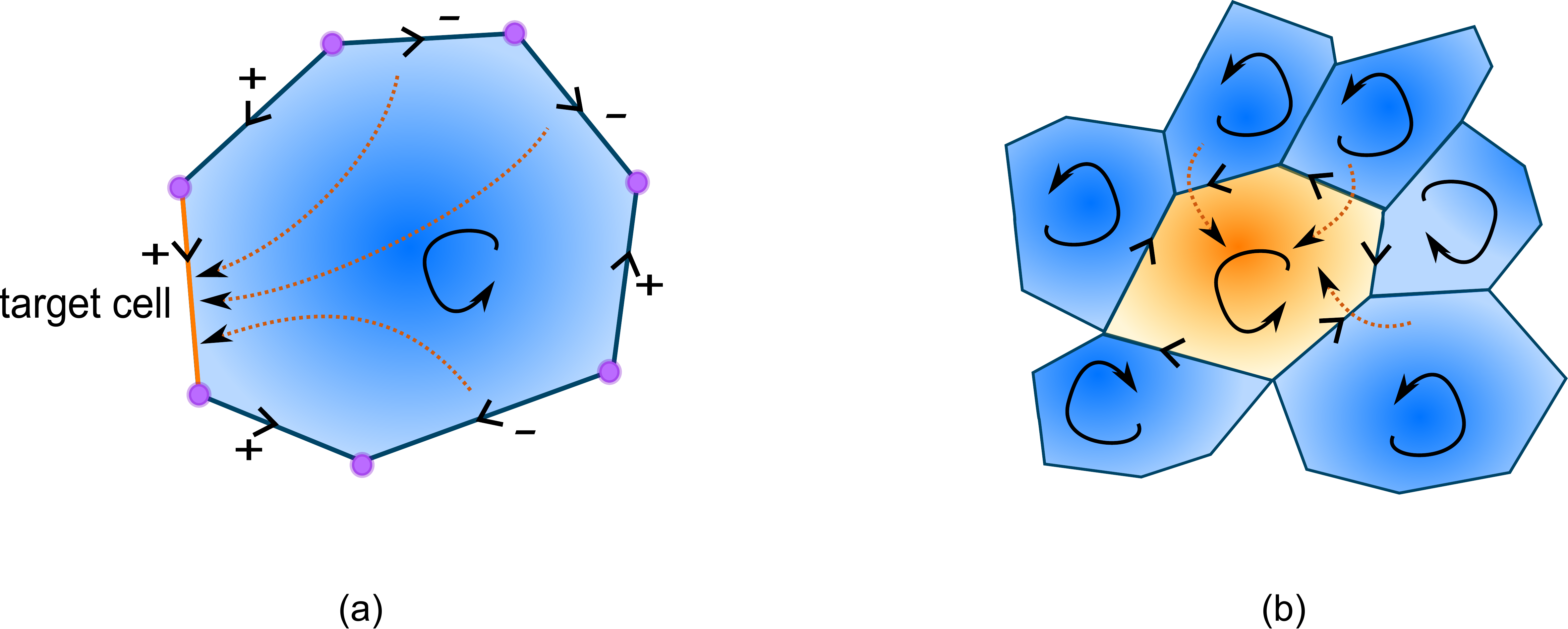}
    \caption{The adjacency and coadjacency of a cell. (a) The adjacent 1-cells of the orange target 1-cells. (b) The coadjacent 2-cells of the orange 2-cell. This adjacency/coadjacency relations will be used for inter-cellular message passing schemes.  }
  \label{example2}}
\end{figure}

In Figure \ref{example} (a), we have $\mathcal{CO}[v_2, v_3 ]=\{e_3,e_4\}$ whereas $\mathcal{CO}[v_3, v_2 ]=\{e_4\}$. On the other hand, we have $\mathcal{CO}[e_1, e_2 ]=\mathcal{CO}[e_2, e_1 ]=\emptyset$ in Figure \ref{example} (b). See also Figure \ref{example2} for an illustrative example on adjacency and coadjacency relationships.

\subsection{General Message Passing Scheme}
\label{general}
The inter-cellular message passing scheme given in Section \ref{cell complex nets} updates the vectors on the flows from a given $0$-cell and gathers the information from surrounding cells in a radial fashion defined by the adjacency matrices of the cell complex. Although this message passing scheme is natural from the perspective of generalizing GNNs passing schemes, it forms a single method out of many other natural methods that can be defined in the context of CXNs. 

\subsubsection{Co-adjacency Message Passing Scheme}
The message passing scheme given in Section \ref{cell complex nets} does not update the vectors associated with the final $n-$ cells on the complex. In certain applications, it might be desirable to make the flow of information go from the lower cells to the higher cells. An example of such an application is mesh segmentation where it is desirable to update the information associated with a face on the mesh after gathering local surrounding information. 
The scheme given in Section \ref{cell complex nets} can be easily adjusted for this purpose.  To this end, we utilize the data on the cells complex as before while re-defining the message passing schemes as follows:

            \begin{equation}
             \small{
         h_{c^{n}}^{ (k) }:= \alpha^{(k)}_{n-1} \bigg( h_{c^{n}}^{ (k-1) }, E_{ a^{n} \in \mathcal{N}_{co}(c^{n})}\Big( \phi^{(k)}_{n-1} ( h_{c^{n}}^{(k-1)},h_{a^{n}}^{(k-1)}, F_{e^{n-1} \in \mathcal{C} [a^{n},c^{n}]  } ( h_{e^{n-1}}^{(k-1))}    ) \Big) \bigg)  \in \mathbb{R}^{l^k_{n}},  }
      \end{equation}
      $\vdots$
        \begin{equation}
         \small{
         h_{c^1}^{ (k) }:= \alpha^{(k)}_1 \bigg( h_{c^1}^{ (k-1) }, E_{ a^1 \in \mathcal{N}_{co}(c^1)}\Big( \phi^{(k)}_1 ( h_{c^1}^{(k-1)},h_{a^1}^{(k-1)}, F_{e^0 \in \mathcal{C}[a^1,c^1]  } ( h_{e^0}^{(0)}    ) \Big) \bigg)  \in \mathbb{R}^{l^k_1}}
               \end{equation}

\begin{figure}[t]
  \centering
   {\includegraphics[scale=0.055]{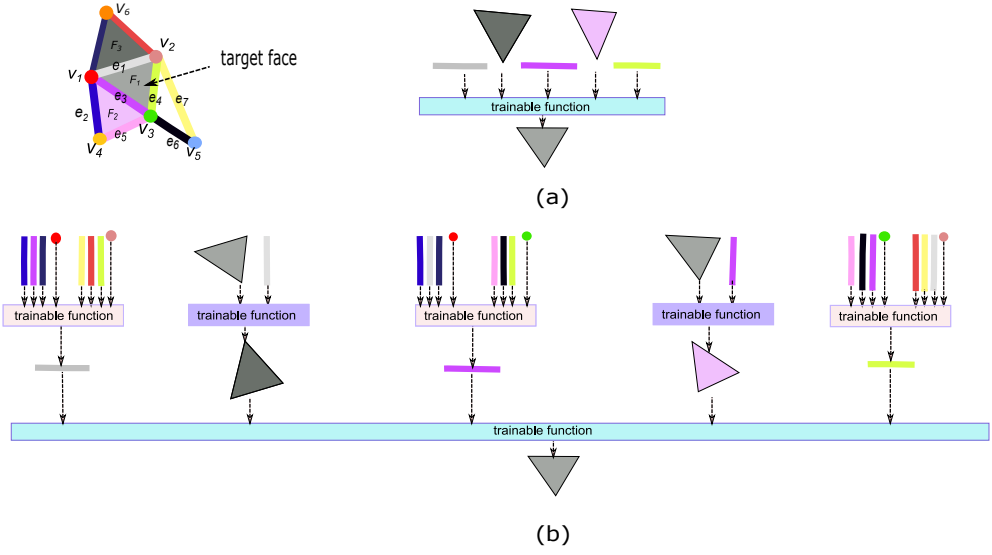}
    \caption{CXN with 2 layers. This example demonstrates a simplicial complex neural network for clarity. The computation is only demonstrated with respect to the light grey face. The information flow goes from higher cells to lower incident cells. }
  \label{DMF123}}
\end{figure}

Note that the initial vector associated with zero cells in $X$ is never updated in this case. An example of these computations is illustrated in Figure \ref{DMF123}.

\subsubsection{Homology and Cohomology-Based Passing Schemes}
This subsection briefly outlines a message passing scheme that is consistent with the boundary and coboundary maps in the context of homology and cohomology of a cell complex. 

Let $h^k$ be a cell in a, possible oriented, cell complex $X$. Let $Bd(x)$ be set of cells $y$ of dimension $k-1$ such that $y\in \partial(x)$ and $x$ and $y$ have compatible orientations. Similarly, let $CoBd(x)$ denotes all cells of $y \in X$ with $h\in \partial(y)$ and both $x$ and $y$ have compatible orientations. Denoted by $\mathcal{I}(x)$ to the union $Bd(x) \cup CoBd(x) $, we may define the passing scheme as follows:  
\begin{equation}
          \small{h_{c^m}^{ (k) }:= \alpha^{(k)}_m \bigg( h_{c^m}^{ (k-1) }, E_{ a \in \mathcal{I}(x)}\Big( \phi^{(k)}_{m,d(a)} ( h_{c^m}^{(k-1)},h_{a}^{(k-1)} ) \Big) \bigg)  \in \mathbb{R}^{l^k_m}
          }
          \end{equation}
        Notice that the trainable function $\phi^{(k)}_{m,d(a)}$ needs to accommodate for the fact that the dimension of the vector associated with $a$, namely $x_{a}^{(k-1)}$, may vary for different $a\in \mathcal{I}(x)$.



\subsection{Potential Applications}
\label{potenial}
The proposed CXNs has several potential applications. For example:
\begin{enumerate}
    \item \textbf{Studying the type of underlying spaces.} The \textit{topological type} of the underlying space is a central question in topology. Specifically, given two spaces $A$ and $B$, are they equivalent up to a given topological equivalence? In practice, this can be translated to a similarity question between two structures. Indeed, TDA has been extensively utilized towards this purpose \cite{dey2010persistent,hajij2018visual}. On the other hand, deep learning allows studying the structure of the underlying space by building complex relationship between various elements in this space. Cell complexes form a general class of topological spaces that encompasses graphs, simplicial complexes, and polyhedral complexes. Hence, CXNs provides a potential tool to study the structure similarity between discrete domains such as 3D shapes and discrete manifolds \footnote{Within this context, GNNs with their current neighborhood aggregation scheme have been shown to not being able to solve the graph isomophism problem \cite{xu2018powerful}.}.
    
    \item \textbf{Learning cell complex representation.} The cell complex autoencoder framework introduced in \ref{autoencoder} extends the applications of graphs autoencoder to a much wider set of possibilities. In particular, cell complexes are natural objects for \textit{language embedding} as they can be used to build complex relationships of arbitrary length. Specifically, we can build a cell complex out of a corpus of text: words are vertices, they share an 1-cells if they are adjacent in the corpus, within a sentence, sentences form the boundaries of 3-cells, paragraphs form the boundaries of 4-cells, chapters form the boundaries of 4-cells, etc. Notice that unlike less general complexes (e.g. simplical complexes), a $k$-cell in a cell complex may have an arbitrary number of $(k-1)$ incident cells.  

\end{enumerate}

\end{document}